\documentclass{article} 
\usepackage{iclr2027_conference,times}

\usepackage{amsmath,amsfonts,bm}

\def\eqref#1{equation~\ref{#1}}

\def\1{\bm{1}}

\DeclareMathAlphabet{\mathsfit}{\encodingdefault}{\sfdefault}{m}{sl}
\SetMathAlphabet{\mathsfit}{bold}{\encodingdefault}{\sfdefault}{bx}{n}

 \usepackage{fontawesome7}
 \usepackage{enumitem}
 \usepackage{wrapfig}
\usepackage[utf8]{inputenc} 
\usepackage{titletoc}
\usepackage[T1]{fontenc}    
\usepackage{url}            
\usepackage{subfigure}
\usepackage{etoc}

\usepackage{fontawesome7}
 \usepackage{enumitem}
 \usepackage{mdframed}
 \usepackage{amsmath}
 \usepackage{multirow} 
 \usepackage{wrapfig}

 \usepackage{listings}

\usepackage{booktabs}       
\usepackage{amsfonts}       
\usepackage{nicefrac}       
\usepackage{pifont}
\usepackage{microtype}      
\usepackage{xcolor}         
\usepackage{algorithm}
\usepackage{algorithmic}
\usepackage{graphicx}
\usepackage{color, colortbl}
\usepackage{soul} 
\usepackage[colorlinks=true, linkcolor=cyan, citecolor=cyan, urlcolor=cyan]{hyperref}

\usepackage{tcolorbox}
\tcbuselibrary{most}
\newtcolorbox{promptbox}[1]{
    enhanced,
    breakable,
    colback=black!2,
    colframe=black!25,
    boxrule=0.6pt,
    arc=1.5mm,
    left=2mm,
    right=2mm,
    top=2mm,
    bottom=1.5mm,
    fonttitle=\small\bfseries\sffamily,
    coltitle=black,
    colbacktitle=black!7,
    title={#1},
    attach boxed title to top left={
        xshift=2mm,
        yshift=-2mm
    },
    boxed title style={
        boxrule=0pt,
        arc=1mm,
        left=1.5mm,
        right=1.5mm,
        top=0.5mm,
        bottom=0.5mm
    }
}

\title{Auditing Privacy Risks in LLM-Enhanced\\Graph Neural Networks}

\author{Longzhu He\textsuperscript{*}\\BUPT
\And
Zelang Wen\textsuperscript{*}\\BUPT
\And
Chaozhuo Li\\BUPT
\And
Sen Su\\BUPT
}

\iclrfinalcopy 
\begin{document}

\maketitle

\begin{abstract}
Large language models (LLMs) have recently advanced graph neural networks (GNNs) by enriching node representations with semantic information, giving rise to \textit{LLM-enhanced GNNs} that achieve substantial performance gains. However, how such semantic enhancement affects privacy risks remains largely underexplored. To bridge this gap, we systematically audit the privacy risks of LLM-enhanced GNNs through a unified framework consisting of five stages: \ding{172} dataset preparation, \ding{173} victim model training, \ding{174} privacy attack, \ding{175} risk assessment, and \ding{176} defense analysis. Specifically, our evaluation spans ten text-attributed graph datasets across diverse domains, six privacy attacks, 42 LLM-enhanced GNN configurations, and three more recent language-model backbones. Extensive experiments show that, despite their utility improvements, LLM-enhanced GNNs consistently exhibit greater empirical privacy vulnerability than shallow text representation baselines under the evaluated attacks across diverse models and datasets. Further analysis shows that LLM-enhanced representations exhibit more distinguishable link-, label-, and membership-related signals in the embedding space, making them more exploitable by inference attacks. Finally, we evaluate representative defenses and examine their effectiveness in mitigating these privacy risks. Overall, this work provides a systematic audit of privacy risks in LLM-enhanced GNNs and offers insights for developing more secure and trustworthy graph learning systems.
\end{abstract}

\section{Introduction}
\vspace{-0.5em}
In recent years, graph neural networks (GNNs)~\citep{wu2020comprehensive,DBLP:conf/iclr/KipfW17} have achieved remarkable success in graph representation learning and have become the dominant paradigm for modeling graph-structured data. In many real-world applications, graphs are not only characterized by their topology but also associated with rich textual attributes, such as user profiles, paper abstracts, or semantic descriptions of entities, forming so-called text-attributed graphs~\citep{yan2023comprehensive,yao2019graph}. Traditional approaches typically rely on shallow text encoding techniques, such as Bag-of-Words (BoW)~\citep{harris1954distributional}, Word2Vec~\citep{church2017word2vec}, or skip-gram models~\citep{mikolov2013distributed}, to obtain initial node representations, which are then integrated into GNN-based message passing frameworks. However, these methods are limited in their ability to capture rich contextual and semantic information in node texts. Consequently, effectively modeling textual semantics while preserving structural dependencies remains a key challenge in graph representation learning.

\begin{figure}[t]
  \centering
\includegraphics[width=\linewidth]{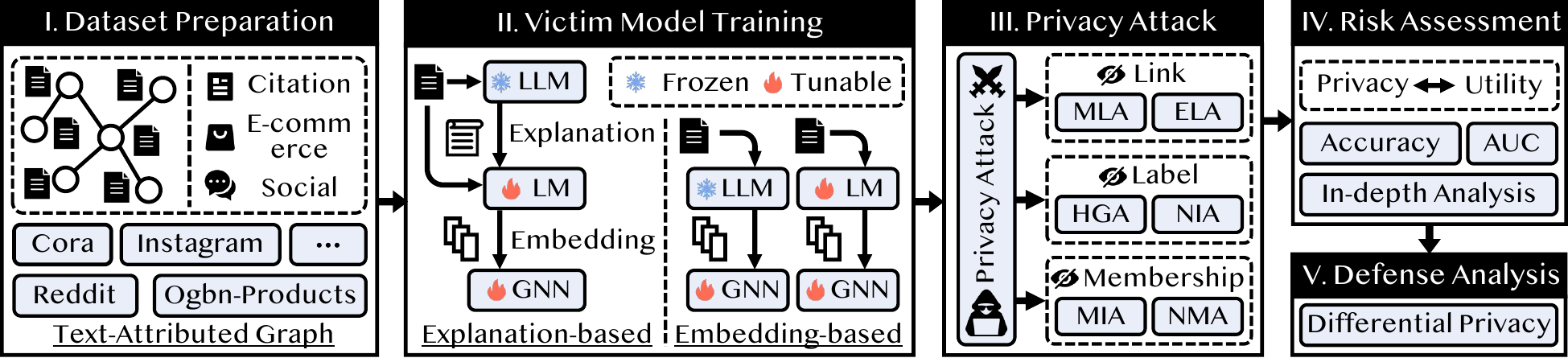}
\vspace{-1.8em}
  \caption{\textbf{Overview of the proposed privacy auditing framework.} It consists of five stages: dataset preparation, victim model training, privacy attack, risk assessment, and defense analysis.}
\label{fig1}
\vspace{-1em}
\end{figure}

Recently, large language models (LLMs) have demonstrated strong capabilities in natural language understanding~\citep{kuang2025natural,yi2025survey}, contextual reasoning~\citep{xiong2025mapping,chi2024unveiling}, and instruction following~\citep{lou2024large,qin2024infobench}, and have been increasingly introduced into graph learning to enhance the semantic modeling of textual attributes. Compared with traditional text encoders, LLMs can generate more expressive and context-aware representations by capturing deep semantic relationships beyond surface-level word co-occurrence patterns. This has led to the emergence of \textit{LLM-enhanced GNNs}~\citep{DBLP:conf/iclr/HeB0PLH24,chen2024exploring,zhu2024efficient,DBLP:conf/iclr/ChienCHYZMD22,zhang2025toward}, a new paradigm that integrates LLMs into graph learning to enrich node representations. As illustrated in Fig.~\ref{fig1}\textcolor{cyan}{(II)}, existing approaches can be broadly categorized into \textit{explanation-based} and \textit{embedding-based} approaches. These approaches typically utilize LLMs or smaller pretrained language models (LMs) to perform semantic augmentation, such as contextual enrichment or representation refinement of node texts, and seamlessly integrate the resulting embeddings into GNN pipelines. By jointly modeling structural dependencies and enriched semantic representations, LLM-enhanced GNNs have achieved substantial performance improvements across various graph learning tasks, such as node classification~\citep{DBLP:conf/iclr/HeB0PLH24}.

Despite their strong empirical performance, the privacy risks of LLM-enhanced GNNs remain largely underexplored. Existing studies~\citep{meng2023devil,zhang2022inference,he2021stealing} have shown that GNNs are highly vulnerable to privacy attacks, where adversaries can infer sensitive information from model outputs, even when only limited access is available. In LLM-enhanced GNNs, such risks may be further exacerbated, as node representations integrate structural dependencies with rich semantic features encoded by LLMs or LMs, potentially increasing the information content exposed through model outputs. This paradigm shift may introduce new and more powerful avenues for privacy leakage beyond those observed in conventional GNNs. Therefore, how privacy vulnerabilities evolve under LLM-enhanced architectures remains poorly understood, calling for a systematic audit of the privacy risks associated with this emerging graph learning paradigm.

To fill this gap, we systematically audit the privacy risks of LLM-enhanced GNNs. As illustrated in Fig.~\ref{fig1}, we develop a unified privacy auditing framework encompassing five stages, including \ding{172} dataset preparation, \ding{173} victim model training, \ding{174} privacy attack, \ding{175} risk assessment, and \ding{176} defense analysis, enabling an end-to-end characterization of privacy vulnerabilities and mitigation strategies.

Specifically, our evaluation spans ten real-world text-attributed graph datasets across diverse domains, with six used in the main benchmark and four for additional cross-domain validation. We evaluate 42 LLM-enhanced GNN configurations across multiple feature enhancers and widely used GNN backbones, together with three more recent language-model backbones, covering diverse model architectures and representations. We then consider six representative privacy attacks targeting three fundamental objectives, namely \textit{link}, \textit{label}, and \textit{membership} inference. These attack objectives have been extensively studied in prior 
work~\citep{olatunji2021membership,meng2023devil,zhang2022inference,he2021stealing,guan2025attention,zhang2024survey} 
on traditional GNNs and are widely recognized as the most prevalent forms of 
privacy leakage, as they respectively capture the exposure of graph structure, 
node semantics, and training participation. In the risk assessment stage, we conduct a quantitative evaluation of privacy risks and further investigate how semantic enhancement affects model vulnerability. The results show that LLM-enhanced representations exhibit stronger structural-, label-, and membership-related signals, providing a plausible explanation for their higher empirical privacy vulnerability. Finally, we evaluate representative differential privacy (DP)~\citep{sajadmanesh2021locally,lin2022towards,zhu2023blink,he2025going,dwork2008differential} defenses covering both input-level perturbation and private GNN training, and characterize their effectiveness and utility costs under different protection settings. Our contributions can be summarized as follows:
\vspace{-0.5em}
\begin{itemize}[leftmargin=*, itemindent=0em]
\renewcommand{\labelitemi}{$\diamond$}
\item \textbf{Important But Underexplored Problem.}
We identify and systematically investigate the important yet underexplored privacy risks in LLM-enhanced GNNs, which have received limited attention in existing graph learning and privacy research.

\item \textbf{Systematic Auditing Framework.}
We develop a unified five-stage privacy auditing framework to systematically evaluate privacy risks in LLM-enhanced GNNs, covering real-world datasets, diverse privacy attacks, in-depth risk analysis, and representative defenses, thereby enabling a holistic understanding of privacy vulnerabilities and mitigation effectiveness across settings.

\item \textbf{Comprehensive Empirical Evaluation.}
Our evaluation spans ten text-attributed graph datasets, six privacy attacks, 42 LLM-enhanced GNN configurations, and three more recent language-model backbones, together with extensive analyses of representative privacy defenses.
\end{itemize}

\section{Preliminaries}
\label{gen_inst}

\vspace{-0.5em}

In this section, we first present the problem definition ($\triangleright$ Sec.~\ref{sec2.1}), and then introduce the necessary background on graph neural networks ($\triangleright$ Sec.~\ref{sec2.2}) and LLM-enhanced GNNs ($\triangleright$ Sec.~\ref{sec2.3}).

\vspace{-0.7em}

\subsection{Problem Definition}\label{sec2.1} 

\vspace{-0.5em}

Consider a text-attributed graph (TAG) $\mathcal{G} = (\mathcal{V}, \mathcal{E}, \mathcal{T}, \mathbf{A})$, where $\mathcal{V}=\{v_1,\cdots,v_N\}$ denotes the set of nodes, $\mathcal{E}$ represents the set of edges, $\mathcal{T} \hspace{-0.3em}=\hspace{-0.3em}\{t_v\}_{v \in \mathcal{V}}$ contains the textual attributes of the nodes, and $\mathbf{A}\in\mathbb{R}^{N\times N}$ is the adjacency matrix with $\mathbf{A}_{ij} \in \{0,1\}$ indicating the presence of an edge between nodes $v_i$ and $v_j$. TAGs seamlessly integrate structured topological information with unstructured textual semantics, providing a versatile representation for a wide range of real-world networks, including social networks, citation networks, and e-commerce networks. Consistent with prior work~\citep{DBLP:conf/iclr/HeB0PLH24,chen2024exploring,zhu2024efficient,DBLP:conf/iclr/ChienCHYZMD22,zhang2025toward}, we focus on the node classification task. The node set $\mathcal{V} = \mathcal{V}_L \cup \mathcal{V}_U$ consists of labeled nodes $\mathcal{V}_L$ and unlabeled nodes $\mathcal{V}_U$, where each labeled node $v \in \mathcal{V}_L$ is associated with a class label $y_v \in \mathcal{Y} = \{1,\dots,C\}$. The objective is to learn a mapping $f_\theta: (\mathbf{A}, \mathcal{T}) \rightarrow \mathcal{Y}$ that predicts labels for unlabeled nodes by jointly exploiting graph structure and textual node attributes. Building on this formulation, we investigate how an adversary can exploit the outputs of LLM-enhanced GNNs, together with their prior knowledge and capabilities, to infer sensitive information about target nodes, including link existence, node labels, and training membership (see Sec.~\ref{sec3} for the threat model).
\vspace{-0.7em}

\subsection{Graph Neural Networks}\label{sec2.2}

\vspace{-0.5em}

GNNs are a class of deep learning models designed for graph-structured data, whose core idea is to learn node representations (\textit{i.e.}, embeddings) by aggregating information from neighboring nodes~\citep{wu2020comprehensive}. Taking advantage of this message passing mechanism, GNNs have achieved strong performance across a wide range of real-world applications, including social network analysis~\citep{DBLP:conf/iclr/KipfW17}, knowledge graph completion~\citep{liu2021indigo}, fraud detection~\citep{yang2025flag}, and recommender systems~\citep{wu2022graph}. Representative architectures include GCN~\citep{DBLP:conf/iclr/KipfW17}, GAT~\citep{DBLP:conf/iclr/VelickovicCCRLB18}, and SAGE~\citep{hamilton2017inductive}.
Formally, let $\mathcal{N}(v)$ denote the set of neighbors of node $v$. At the $k$-th layer, node representations are computed via a two-step procedure consisting of neighborhood aggregation and representation update:
\begin{equation}
\mathbf{h}^{k}_{\mathcal{N}(v)} = \textsc{Aggregate}_k(\left\{\mathbf{h}^{k-1}_u \mid u \in \mathcal{N}(v)\right\}),\quad
\mathbf{h}^{k}_v = \textsc{Update}_k(\mathbf{h}_{v}^{(k-1)},\mathbf{h}^{k}_{\mathcal{N}(v)}),     
\end{equation}
where $\mathbf{h}^{k-1}_u$ represents the representation of neighbor $u$ at layer $k-1$. The operator $\textsc{Aggregate}_k(\cdot)$ performs permutation-invariant aggregation over neighbors, such as mean, sum, or max pooling, while $\textsc{Update}_k(\cdot)$ denotes a learnable non-linear transformation, typically parameterized by a neural network. At the input layer, $\mathbf{h}^{0}_v$ is initialized with the original feature vector $\mathbf{x}_v$ of node $v$.

\vspace{-0.7em}

\subsection{LLM-Enhanced GNNs}\label{sec2.3} 

\vspace{-0.5em}

With the rapid advancement of LLMs, recent studies~\citep{DBLP:conf/iclr/HeB0PLH24,chen2024exploring,zhu2024efficient,DBLP:conf/iclr/ChienCHYZMD22} have explored integrating them with GNNs, giving rise to \textit{LLM-enhanced GNNs}\footnote{Following prior work~\citep{DBLP:conf/ijcai/0001LW0S0Y24,DBLP:conf/iclr/HeB0PLH24,chen2024exploring,ma2026llm,zhang2025toward,yang2025flag}, we use \textit{LLM-enhanced GNNs} as a broad term encompassing GNNs augmented by either large language models (LLMs) or pretrained language models (LMs).}. These approaches leverage LLMs to encode textual node attributes, introducing rich semantic information that complements the limited expressiveness of traditional GNNs on unstructured text. As a result, they can significantly improve node representations and downstream task performance. Depending on whether additional textual information is generated by LLMs, as illustrated in Fig.~\ref{fig1}\textcolor{cyan}{(II)}, existing methods can be broadly categorized into two groups: 

\vspace{-0.5em}

\begin{itemize}[leftmargin=*, itemindent=0em]
\renewcommand{\labelitemi}{$\diamond$}
\item \textbf{Explanation-based methods} exploit the reasoning and knowledge capabilities of LLMs to derive high-level semantic information (\textit{e.g.}, explanations) from raw textual attributes, which is then encoded into node embeddings by an LM (\textit{e.g.}, DeBERTa-base~\citep{DBLP:conf/iclr/HeLGC21}). Formally, this process can be written as $\mathbf{x}_v = f_{\text{LM}}(f_{\text{LLM}}(p, t_v), t_v)$,  where $f_{\text{LLM}}$ generates augmented text conditioned on a prompt $p$, and $f_{\text{LM}}$ encodes both original and augmented text into embeddings.
\item \textbf{Embedding-based methods} do not explicitly generate additional text but instead directly encode textual attributes into embeddings. These methods can be further divided into two variants: \ding{172} directly using LLMs to obtain embeddings, \textit{i.e.}, $\mathbf{x}_v = f_{\text{LLM}}(t_v)$, and \ding{173} fine-tuning conventional LMs to produce task-specific embeddings, \textit{i.e.}, $x_v = f_{\text{LM}}(t_v)$. 
Such approaches typically rely on embedding-accessible LLMs or fine-tunable pretrained models to adapt to downstream tasks.
\end{itemize}

\vspace{-1em}

\section{Threat Model}\label{sec3}

\vspace{-0.5em}

We characterize the threat model in terms of the \textit{knowledge}, \textit{capability}, and \textit{objective} of the adversary.

\begin{wrapfigure}{r}{0.4\linewidth}
\vspace{-1.5em}
  \centering
\includegraphics[width=\linewidth]{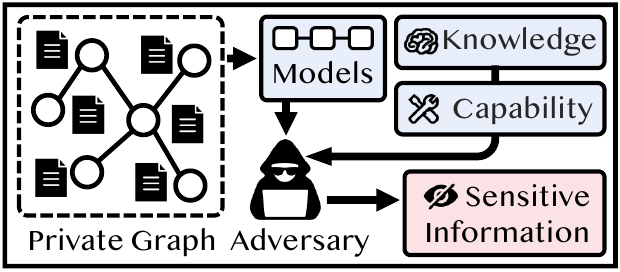}
\vspace{-1.7em}
  \caption{Adversary–model interaction illustration under black-box access.}
\label{fig2}
\vspace{-1em}
\end{wrapfigure}

\textbf{Adversary’s Knowledge and Capability.} We consider an adversary with \textit{black-box access} to the target GNN model. Specifically, the adversary has no access to the model’s internal parameters or architecture and can only obtain node posteriors through query access. This setting is consistent with prior attack literature~\citep{meng2023devil,zhang2022inference,he2021stealing,fu2025safeguarding} and aligns with real-world deployments of \textit{machine-learning-as-a-service} (MLaaS) platforms, where models are exposed via query interfaces, as exemplified by Google Cloud’s Vertex AI~\citep{cite1}, Meta Platforms’ ParlAI~\citep{cite3}, and IBM’s InfoSphere Virtual Data Pipeline~\citep{cite4}. Beyond query access, the adversary may possess auxiliary prior knowledge $\mathcal{K}$ and additional capabilities to facilitate the attack. As illustrated in Fig.~\ref{fig2}, the adversary interacts with the model to collect posterior predictions, which are then exploited to infer sensitive information about the underlying graph data.

\textbf{Adversary’s Objective.} We focus on three fundamental privacy risks in graph learning, namely \textit{link}, \textit{label}, and \textit{membership} inference. These attack objectives are widely studied in prior work~\citep{olatunji2021membership,guan2025attention,meng2023devil,zhang2022inference,wu2022linkteller,he2021stealing,zhang2024survey} on traditional GNNs and are considered representative forms of privacy leakage, as they respectively target graph structure, node semantics, and training participation. By adopting this taxonomy, our evaluation remains consistent with established attack paradigms while enabling systematic assessment of privacy risks in LLM-enhanced GNNs. Formally, given a target node $v$ and its model posterior $\mathcal{R}_v$, the adversary’s objectives are defined as follows:

\vspace{-0.5em}

\begin{itemize}[leftmargin=*, itemindent=0em]
\renewcommand{\labelitemi}{$\diamond$}
\item \textbf{Link Inference.} The adversary attempts to infer whether a link exists between node $v$ and another node $u$. This reveals structural relationships in the graph, which may correspond to sensitive interactions (\textit{e.g.}, social ties). The objective is: $\hat{e}_{vu}
=
\arg\max_{e\in\{0,1\}}
\Pr(e_{vu}=e\mid \mathcal{R}_v,\mathcal{R}_u).$
\item \textbf{Label Inference.} The adversary aims to infer the true class label $y_v$ of node $v$. The label $y_v$ can expose sensitive information about node $v$, \textit{e.g.}, personalized recommendations that reveal their private interests. The formal objective is defined as: $\hat{y}_v
=
\arg\max_{c\in\mathcal{Y}}
\Pr(y_v=c\mid \mathcal{R}_v)$.
\item \textbf{Membership Inference.} The adversary aims to determine whether node $v$ was included in the training set of the target model. Such information can reveal participation in sensitive datasets (\textit{e.g.}, medical or user-specific records). The objective is: $\hat{m}_v
=
\arg\max_{m\in\{0,1\}}
\Pr(m_v=m\mid \mathcal{R}_v)$, where $m_v \in \{0,1\}$ indicates whether $v$ is a member of the training data.
\end{itemize}

\vspace{-0.5em}

\vspace{-0.7em}

\section{Privacy Auditing Framework}
\label{sec4}

\vspace{-0.7em}

In this section, we present the five stages of our privacy auditing framework in sequence: \textit{dataset preparation} ($\triangleright$ Sec.~\ref{sec4.1}), \textit{victim model training} ($\triangleright$ Sec.~\ref{sec4.2}), \textit{privacy attack} ($\triangleright$ Sec.~\ref{sec4.3}), \textit{risk assessment} ($\triangleright$ Sec.~\ref{sec4.4}), and \textit{defense analysis} ($\triangleright$ Sec.~\ref{sec4.5}). Fig.~\ref{fig1} illustrates the overall workflow.

\begin{wraptable}{r}{0.59\textwidth}
\vspace{-4.5em}
\setlength\tabcolsep{4.5pt}
\small
\caption{Statistics of datasets.}\label{tab:abl_key_module}
\vspace{4pt}
\label{tab1}
\centering
\begin{tabular}{l|rrrr}
    \toprule
      \textbf{Dataset} & \textbf{\#Nodes} & \textbf{\#Edges}  & \textbf{\#Classes}  & \textbf{Domain}\\
    \midrule
    Cora & 2,708 & 5,429 & 7 & Citation \\
     CiteSeer & 3,327 & 4,552 & 6 & Citation \\
    Ogbn-Products & 12,011 &21,987 & 47 & E-commerce  \\
     Tape-Arxiv23  & 13,167 &23,735 & 40 & Citation  \\
     Instagram & 11,339 & 144,010 & 2 & Social \\
     Reddit & 33,434 & 198,448 & 2 & Social \\
    \bottomrule
    \end{tabular}
\vspace{-10pt}
\end{wraptable}

\vspace{-0.5em}

\subsection{Dataset Preparation}\label{sec4.1}

\vspace{-0.5em}

We consider six TAG datasets in the main evaluation, with statistics summarized in Table~\ref{tab1}. They span three real-world scenarios: social networks, citation networks, and e-commerce platforms. The social networks include \textit{Instagram}~\citep{huang2024can} and \textit{Reddit}~\citep{huang2024can}, where nodes represent users and textual attributes correspond to user-generated content or profiles. The citation networks include \textit{Cora}~\citep{mccallum2000automating}, \textit{CiteSeer}~\citep{yang2016revisiting}, and \textit{Tape-Arxiv23}  (subset)~\citep{DBLP:conf/iclr/HeB0PLH24}, where nodes denote papers and textual attributes are derived from titles or abstracts. For e-commerce, we adopt \textit{Ogbn-Products}  (subset)~\citep{DBLP:conf/iclr/HeB0PLH24}, where nodes correspond to products with textual descriptions or reviews. These datasets provide diverse structural and textual characteristics for cross-domain privacy evaluation. Further details are provided in Appendix, with four additional datasets evaluated in Appendix.

\vspace{-0.7em}

\subsection{Victim Model Training}\label{sec4.2}

\vspace{-0.5em}

In this stage, we consider 42 victim model configurations for training. Each configuration consists of two key components: an LM/LLM-based feature enhancer and a GNN backbone. To systematically cover different integration paradigms, we select a representative set of feature enhancement methods:

\vspace{-0.5em}

\begin{itemize}[leftmargin=*, itemindent=0em]
\renewcommand{\labelitemi}{$\diamond$}
\item  \textbf{Explanation-based methods.} We consider \textit{TAPE}~\citep{DBLP:conf/iclr/HeB0PLH24} and \textit{KEA}~\citep{chen2024exploring}, two representative approaches that leverage LLMs (\textit{e.g.}, \text{GPT-3.5-Turbo}) to generate complementary semantic information, such as explanations or keyphrases. The generated text is then jointly encoded with the original textual attributes using LMs (\textit{e.g.}, \text{DeBERTa-base}~\citep{DBLP:conf/iclr/HeLGC21} or \text{E5-Large}~\citep{wang2022text}) to produce node embeddings.
\item  \textbf{Embedding-based methods.} As described in Sec.~\ref{sec2.3}, these methods follow two implementation paradigms. The first directly leverages LLMs to generate text embeddings, for which we consider \text{LLaMA-2-7b-hf} (\textit{LLaMA})~\citep{touvron2023llama} and \text{Linq-Embed-Mistral} (\textit{Linq})~\citep{choi2024linq}. The second fine-tunes the LMs to obtain task-specific embeddings; in this category, we include \textit{SimTeg}~\citep{duan2023simteg} and \textit{E5-Large}~\citep{wang2022text}.
\end{itemize}
\vspace{-0.5em}

Based on these feature enhancers, we further combine them with seven GNN backbones: \textit{GCN}~\citep{DBLP:conf/iclr/KipfW17}, \textit{SAGE}~\citep{hamilton2017inductive}, \textit{GAT}~\citep{DBLP:conf/iclr/VelickovicCCRLB18}, \textit{GIN}~\citep{DBLP:conf/iclr/XuHLJ19}, \textit{APPNP}~\citep{DBLP:conf/iclr/KlicperaBG19}, \textit{SGC}~\citep{wu2019simplifying}, and \textit{SSGC}~\citep{DBLP:conf/iclr/ZhuK21}. In total, this results in 42 victim model configurations. Rather than pursuing more advanced architectures, we focus on widely adopted and well-established combinations to ensure reproducibility and provide representative coverage of mainstream LLM-enhanced GNNs. See Appendix for more details. 

\vspace{-0.5em}

\subsection{Privacy Attack}\label{sec4.3}

\vspace{-0.5em}
In this stage, we evaluate privacy risks across three fundamental attack surfaces, namely \textit{link}, \textit{label}, and \textit{membership} inference, covering six representative attack methods in total. These three categories represent prevalent and well-studied privacy attack threats, targeting the exposure of graph structure, node semantics, and training participation, respectively. To ensure a rigorous and comprehensive assessment, the selected attacks span a range of threat model assumptions, from label-aware to fully black-box settings. Brief descriptions are provided below, with full details deferred to Appendix.

\vspace{-0.5em}

\begin{itemize}[leftmargin=*, itemindent=0em]
\renewcommand{\labelitemi}{$\diamond$}

\item \textbf{Link Inference.}
We adopt posterior-based link inference attacks~\citep{he2021stealing} that predict edge existence 
by measuring pairwise distances between node posteriors, using 
\textit{Manhattan distance} (\texttt{MLA}) and \textit{Euclidean distance} (\texttt{ELA}), 
capturing structural leakage from model outputs.

\item \textbf{Label Inference.}
We consider two label inference attacks of different adversarial strength. \textit{Homophily Guessing Attack} (\texttt{HGA})~\citep{meng2023devil} exploits graph homophily~\citep{zhu2020beyond} by 
inferring the victim's label via majority vote over neighbor labels. 
\textit{Node Infiltration Attack} (\texttt{NIA})~\citep{meng2023devil} injects a crafted node connected to the victim and infers its label from model posteriors without accessing label information, representing a complementary threat model.

\item \textbf{Membership Inference.} Two attacks are examined. \texttt{MIA}~\citep{olatunji2021membership} trains a shadow GNN to mimic the target model and a binary MLP on member and non-member posteriors to infer training membership. \texttt{NMA}~\citep{he2021node} extracts top-2 posterior probabilities from 0-hop and 2-hop queries, fuses them through linear layers, and feeds them into a binary MLP classifier.

\end{itemize}

\vspace{-0.7em}
\subsection{Risk Assessment}\label{sec4.4}

\vspace{-0.5em}

In this stage, we adopt standard evaluation metrics to assess the effectiveness of the three types of attacks introduced in Sec.~\ref{sec4.3}. Specifically, for membership inference attacks, we use \textit{attack accuracy} and \textit{AUC} as the primary metrics. For link inference attacks, we adopt \textit{AUC} to measure the ability to distinguish existing and non-existing edges. For label inference attacks, we use \textit{attack accuracy} to evaluate the recovery of node labels. These metrics capture the extent of privacy leakage from different perspectives at the representation level. In addition, we incorporate model utility into the evaluation by using \textit{node classification accuracy} as the utility metric. We also consider distributional measures such as \textit{cumulative distribution functions (CDFs)} of embeddings for more fine-grained analysis. By jointly analyzing privacy risk and utility, we aim to systematically characterize the trade-off between performance improvement and privacy leakage in LLM-enhanced GNNs.

\vspace{-0.7em}

\subsection{Defense Analysis}\label{sec4.5}

\vspace{-0.5em}

In this stage, we evaluate representative differential privacy (DP)~\citep{dwork2008differential} defenses for LLM-enhanced GNNs. DP provides formal privacy guarantees through random noise injection and has been widely adopted in graph learning. Specifically, we consider two complementary privacy settings. \textit{First}, we evaluate several classical local DP (LDP) mechanisms, including 1B~\citep{ding2017collecting}, LP~\citep{phan2017adaptive}, AG~\citep{balle2018improving}, SW~\citep{li2020estimating}, MB~\citep{sajadmanesh2021locally}, PM~\citep{wang2019collecting}, and RR~\citep{warner1965randomized}, which protect sensitive node representations, links, or labels before model training. \textit{Second}, we consider GAP~\citep{sajadmanesh2023gap}, a graph-specific DP method that protects privacy during GNN training. Details are provided in Appendix. Together, these mechanisms allow us to assess privacy defenses across different protection stages and analyze their privacy-utility trade-offs.

\begin{table}[t]
\caption{Node classification accuracy (\%) across different datasets based on the GCN backbone. The highest results are highlighted with \colorbox{blue!20}{\textbf{bold}}, while the second-best results are marked with \colorbox{cyan!15}{\underline{underline}}.}
\vspace{-0.2em}
\centering
\small
\setlength{\tabcolsep}{1.4mm}
\begin{tabular}{l|ccccccc}
\toprule
\textbf{Dataset}
& \textbf{Shallow}
& \textbf{TAPE}
& \textbf{KEA}
& \textbf{LLaMA}
& \textbf{Linq}
& \textbf{SimTeg}
& \textbf{E5-Large}
\\
\midrule
Cora
& 80.8 $\pm$ 1.1
& 82.1 $\pm$ 1.1
& 83.2 $\pm$ 0.8
& 82.2 $\pm$ 0.6
& \cellcolor{blue!20} \textbf{84.8 $\pm$ 0.4}
& 83.3 $\pm$ 0.9
& \cellcolor{cyan!15} \underline{84.5 $\pm$ 0.9}
\\
Tape-Arxiv23
& 55.4 $\pm$ 0.7
& \cellcolor{blue!20} \textbf{72.3 $\pm$ 0.6}
& 65.6 $\pm$ 0.5
& 70.3 $\pm$ 0.7
& \cellcolor{cyan!15} \underline{71.4 $\pm$ 0.4}
& 70.7 $\pm$ 0.6
& 71.3 $\pm$ 0.5
\\
CiteSeer
& 72.3 $\pm$ 1.2
& 73.6 $\pm$ 1.6
& 75.2 $\pm$ 1.5
& 75.3 $\pm$ 1.6
& \cellcolor{blue!20} \textbf{76.6 $\pm$ 0.6}
& 74.9 $\pm$ 0.8
& \cellcolor{cyan!15} \underline{76.2 $\pm$ 2.1}
\\
Ogbn-Products
& 65.6 $\pm$ 0.5
& \cellcolor{blue!20} \textbf{83.1 $\pm$ 0.2}
& 80.9 $\pm$ 0.3
& 81.2 $\pm$ 0.3
& 81.7 $\pm$ 0.8
& 81.4 $\pm$ 0.2
& \cellcolor{cyan!15} \underline{82.2 $\pm$ 0.1}
\\
Instagram
& 63.7 $\pm$ 0.7
& 65.6 $\pm$ 0.8
& 66.1 $\pm$ 0.6
& \cellcolor{cyan!15} \underline{67.3 $\pm$ 0.4}
& \cellcolor{blue!20} \textbf{67.7 $\pm$ 0.7}
& 64.5 $\pm$ 0.7
& 65.6 $\pm$ 0.9
\\
Reddit
& 60.0 $\pm$ 0.9
& 61.5 $\pm$ 1.0
& 61.2 $\pm$ 1.0
& \cellcolor{blue!20} \textbf{62.8 $\pm$ 0.3}
& \cellcolor{cyan!15} \underline{62.5 $\pm$ 1.1}
& 62.3 $\pm$ 0.4
& 61.1 $\pm$ 0.5
\\
\bottomrule
\end{tabular}
\label{tab2}
\vspace{-1em}
\end{table}

\vspace{-1em}

\section{Evaluation}\label{sec5}

\vspace{-0.5em}

In this section, we conduct a comprehensive experimental study to evaluate the privacy risks in LLM-enhanced GNNs. We begin by introducing the experimental setup ($\triangleright$ Sec.~\ref{sec5.1}), followed by detailed results and analyses ($\triangleright$ Sec.~\ref{sec5.2}--Sec.~\ref{sec5.6}), structured around the following key questions:

\vspace{-0.5em}
\begin{itemize}[leftmargin=*, itemindent=0em]
\renewcommand{\labelitemi}{$\diamond$}
\item[$\diamond$] \textbf{RQ1:} How much do LLM-enhanced GNNs improve utility? ($\triangleright$ Sec.~\ref{sec5.2})
    \item[$\diamond$] \textbf{RQ2:} How vulnerable are LLM-enhanced GNNs to link inference attacks? ($\triangleright$ Sec.~\ref{sec5.3})
    \item[$\diamond$] \textbf{RQ3:} How vulnerable are LLM-enhanced GNNs to label inference attacks? ($\triangleright$ Sec.~\ref{sec5.4})
    \item[$\diamond$] \textbf{RQ4:} How vulnerable are LLM-enhanced GNNs to membership inference attacks? ($\triangleright$ Sec.~\ref{sec5.5})
    \item[$\diamond$] \textbf{RQ5:} How effective are defense mechanisms in mitigating these privacy risks? ($\triangleright$ Sec.~\ref{sec5.6})
\end{itemize}

\vspace{-1em}

\subsection{Experimental Settings}\label{sec5.1}

\vspace{-0.5em}

\textbf{Implementation Details.} All experiments are conducted on a machine running Ubuntu 20.04 LTS, equipped with dual Intel\textsuperscript{\textregistered} Xeon\textsuperscript{\textregistered} Gold 6348 CPUs, 100GB RAM, and an NVIDIA\textsuperscript{\textregistered} A800 80GB GPU. GNN models are implemented using PyTorch-Geometric (PyG)\footnote{\url{https://www.pyg.org}}, while LLMs and text encoders are accessed via official APIs or Hugging Face Transformers~\citep{wolf2020transformers}.  Following standard practice~\citep{fu2025safeguarding,ma2026llm}, each dataset is split into training, validation, and test sets with a ratio of 10\%/10\%/80\%. Hyperparameters for GNN models are selected via grid search to ensure competitive node classification performance. Unless otherwise specified, GCN, SAGE, GAT, and GIN use two message-passing layers with a hidden dimension of 256, a dropout rate of 0.5, and a learning rate of 0.001. For GAT, the first layer employs eight attention heads and the second layer uses a single head. APPNP, SGC, and SSGC follow their propagation-specific configurations detailed in Appendix. To reduce randomness, each experiment is repeated five times with different random seeds, and the reported results correspond to the averaged performance.

\textbf{Baseline.} To assess whether LLM/LM-augmented node embeddings introduce additional privacy risks, we adopt shallow embeddings constructed using Bag-of-Words (BoW)~\citep{harris1954distributional} and Word2Vec~\citep{church2017word2vec} as the primary baseline. We additionally include a vanilla pretrained LM, \textit{Pure DeBERTa}~\citep{DBLP:conf/iclr/HeLGC21}, which directly encodes the original node text without LLM-generated augmentation, to more clearly disentangle general semantic richness from LLM-specific enhancement. Details of the \textit{Shallow} baseline and the \textit{Pure DeBERTa} comparison are provided in Appendix.

\vspace{-0.7em}
\subsection{Utility Evaluation}\label{sec5.2}

\vspace{-0.5em}
Before evaluating privacy risks, we first assess whether LLM-enhanced GNNs indeed yield utility gains over shallow text representations. We conduct node classification experiments across all six main datasets, with results reported in Table~\ref{tab2} and full details provided in Appendix. The results show that LLM-enhanced GNNs consistently outperform the \textit{Shallow} baseline in node classification accuracy (\%), confirming that semantic enrichment from LLMs/LMs effectively improves node representation quality. These utility gains motivate a deeper investigation into their privacy implications, as stronger representations may simultaneously expose more sensitive information.

\begin{figure*}[t]
    \centering
    \begin{minipage}{\textwidth}
        \centering
        \subfigure[\normalsize Cora]{\includegraphics[width=0.33\textwidth]{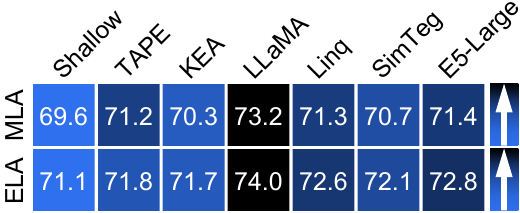}}
        \hspace{-0.1cm}\subfigure[\normalsize Tape-Arxiv23]{\includegraphics[width=0.33\textwidth]{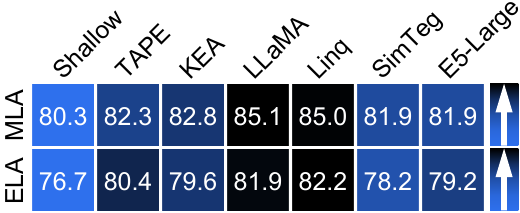}}
        \hspace{-0.1cm}\subfigure[\normalsize CiteSeer]{\includegraphics[width=0.33\textwidth]{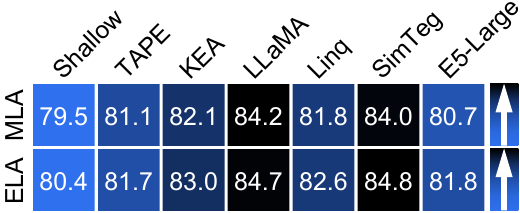}}
     \end{minipage}\\ 
     \vspace{-1em}
\caption{Link inference results (\texttt{MLA}/\texttt{ELA}, AUC \%) on Cora, Tape-Arxiv23, and CiteSeer with GCN. Additional results across more datasets and settings are provided in Appendix.}
    \label{fig:link_main}
    \vspace{-1.6em}
\end{figure*}

\begin{figure*}[t]
    \centering
     \begin{minipage}{\textwidth}
        \centering
        \subfigure[\normalsize Cora]{\includegraphics[width=0.33\textwidth]{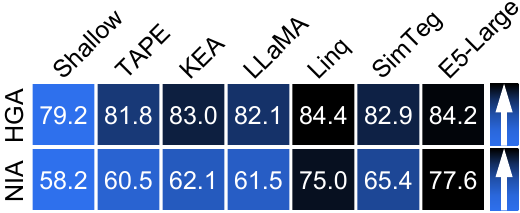}}
        \hspace{-0.1cm}\subfigure[\normalsize Tape-Arxiv23]{\includegraphics[width=0.33\textwidth]{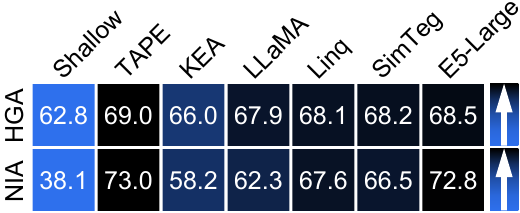}}
        \hspace{-0.1cm}\subfigure[\normalsize CiteSeer]{\includegraphics[width=0.33\textwidth]{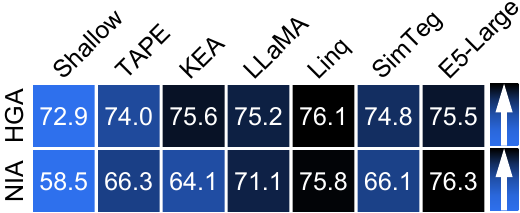}}
     \end{minipage}\\ 
     \vspace{-1em}
   \caption{Label inference results (\texttt{HGA}/\texttt{NIA}, Accuracy \%) on Cora, Tape-Arxiv23, and CiteSeer with GCN. Additional results across more datasets and settings are provided in Appendix.}

\label{fig:label_main}

    \vspace{-1.5em}
\end{figure*}

\begin{figure*}[t]
    \centering
      \begin{minipage}{\textwidth}
        \centering
        \subfigure[\normalsize Cora]{\includegraphics[width=0.33\textwidth]{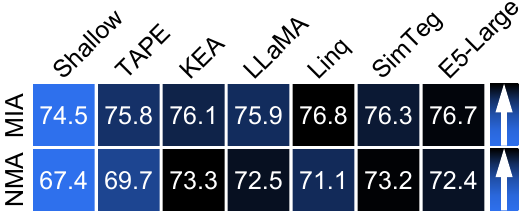}}
        \hspace{-0.1cm}\subfigure[\normalsize Tape-Arxiv23]{\includegraphics[width=0.33\textwidth]{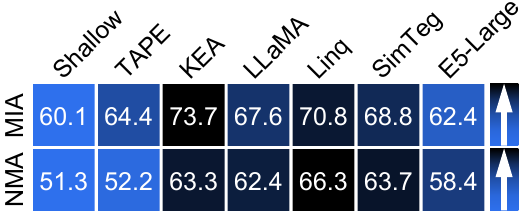}}
        \hspace{-0.1cm}\subfigure[\normalsize CiteSeer]{\includegraphics[width=0.33\textwidth]{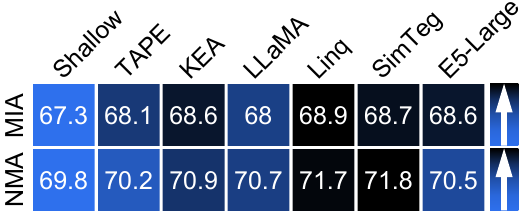}}
     \end{minipage}\\ 
     \vspace{-0.8em}
   \caption{Membership inference results (\texttt{MIA}/\texttt{NMA}, AUC/Accuracy \%) on Cora, Tape-Arxiv23, and CiteSeer with GCN. Additional results across more datasets and settings are provided in Appendix.}
    \label{fig:membership_main}
    \vspace{-1.7em}
\end{figure*}

\vspace{-0.7em}
\subsection{Vulnerability to Link Inference}\label{sec5.3}
\vspace{-0.5em}
We conduct experiments using two link inference attacks (\texttt{MLA} and \texttt{ELA}), with results reported in Fig.~\ref{fig:link_main} and additional results and analyses provided in Appendix. The results reveal two key observations: \ding{172} Compared to the \textit{Shallow} baseline, LLM-enhanced GNNs generally exhibit higher attack AUC (\%), indicating that semantic enrichment increases the vulnerability of node representations to link inference attacks. \ding{173} Across different LLM-enhanced GNN variants, the attack AUC values remain relatively close to one another, suggesting that the increased privacy vulnerability is a general effect of semantic enrichment rather than being driven by any particular model.

\noindent\textbf{In-depth Analysis.} We analyze the distribution of embedding distances between connected and unconnected node pairs on the Tape-Arxiv23 dataset using GCN. As shown in Fig.~\ref{fig7}\textcolor{cyan}{(a)}, compared to \textit{Shallow}, \textit{Linq} exhibits a pronounced separation between connected and unconnected node pairs: distances between connected pairs decrease while those between unconnected pairs increase, resulting in a larger margin in the embedding space. This increased geometric separability provides a plausible explanation for the higher link-inference attack performance.

 \begin{figure*}[t]
    \centering
    \begin{minipage}{\textwidth}
        \centering
        \subfigure[\normalsize Link Inference]{\includegraphics[width=0.33\textwidth]{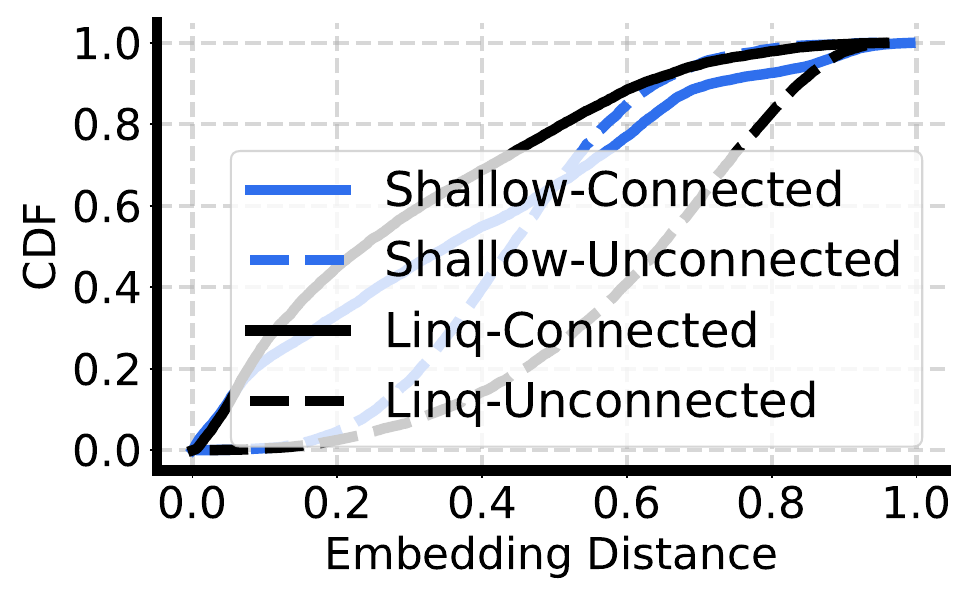}}
        \hspace{-0.15cm}\subfigure[\normalsize Label Inference]{\includegraphics[width=0.33\textwidth]{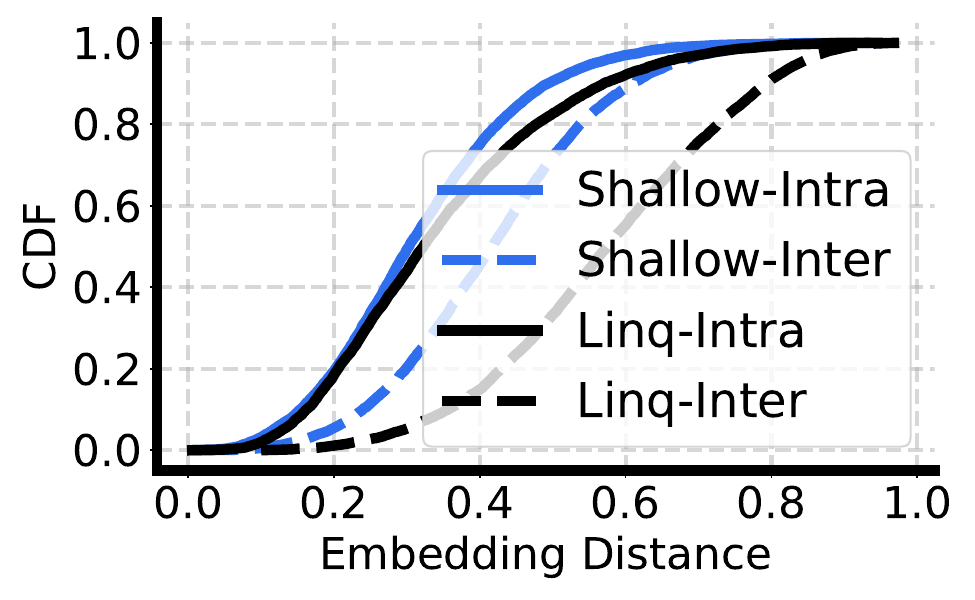}}
\hspace{-0.15cm}\subfigure[\normalsize Membership Inference]{\includegraphics[width=0.33\textwidth]{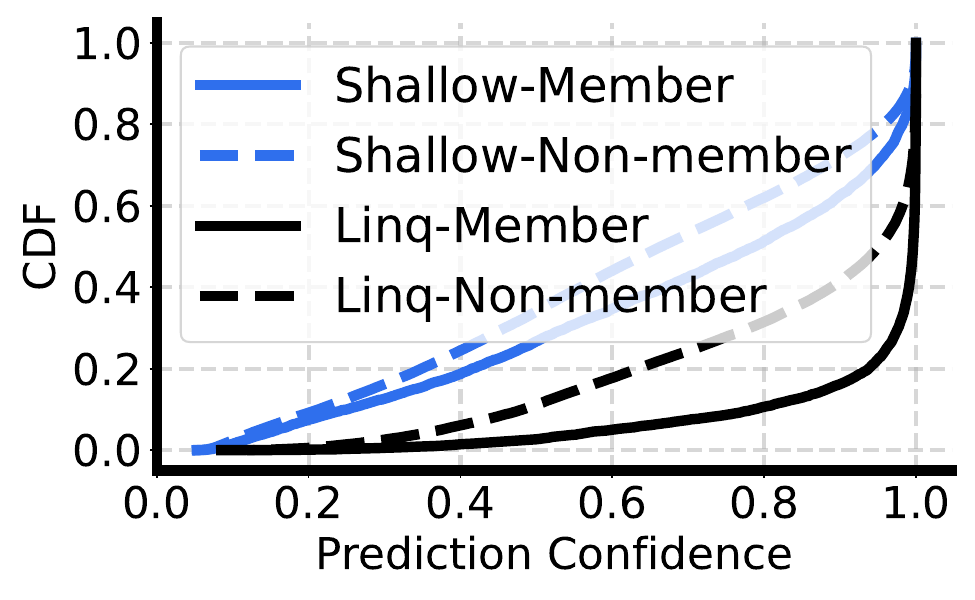}}

     \end{minipage}\\ 
 \vspace{-0.5em}
   \caption{In-depth analysis of privacy vulnerability on Tape-Arxiv23 with GCN, comparing \textit{Linq} against the \textit{Shallow} baseline. (a) Embedding distance distributions of connected \textit{vs.}\ unconnected pairs (link inference). (b) CDF of embedding distances of intra-class\textit{ vs.}\ inter-class pairs (label inference). (c) Prediction confidence distributions of member\textit{ vs.}\ non-member nodes (membership inference).}
\label{fig7}
    \label{fig7}
    \vspace{-1.5em}
\end{figure*}

\begin{figure*}[t]
    \centering
   \begin{minipage}{\textwidth}
        \centering
        \subfigure[\normalsize Link Attack]{\includegraphics[width=0.245\textwidth]{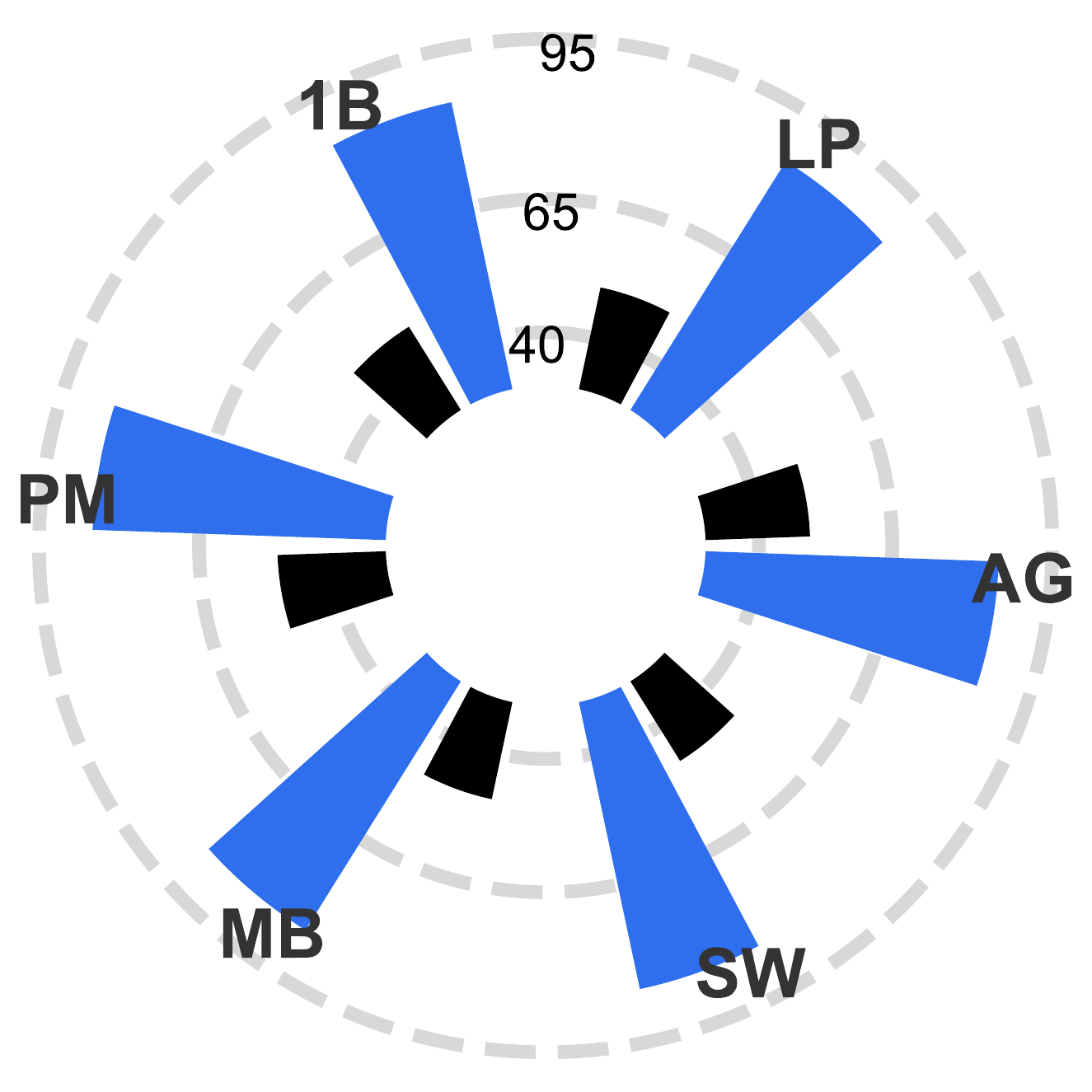}}
        \hspace{-0.1cm}\subfigure[\normalsize Label Attack]{\includegraphics[width=0.245\textwidth]{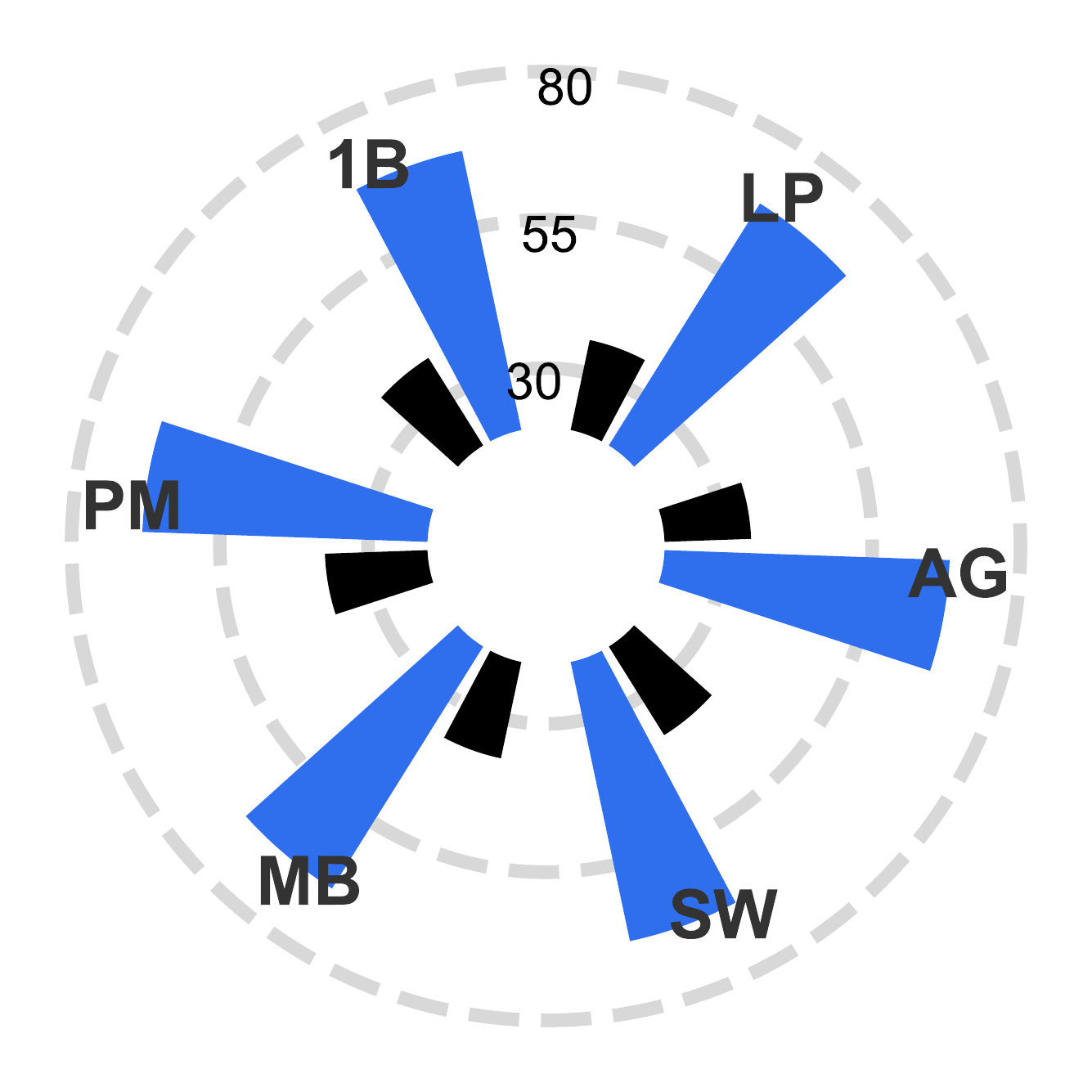}}
        \hspace{-0.1cm}\subfigure[\normalsize Membership Attack]{\includegraphics[width=0.245\textwidth]{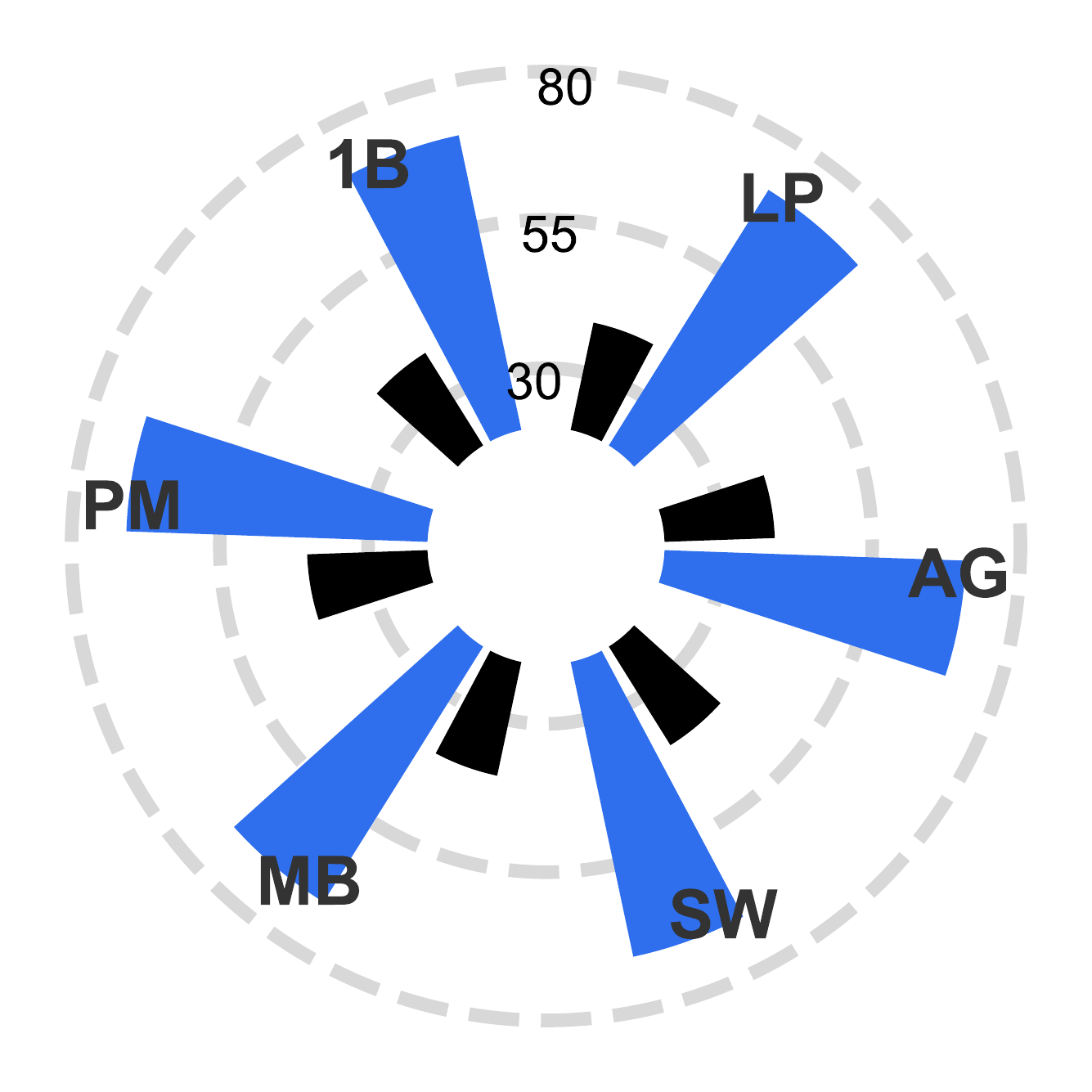}}
         \hspace{-0.1cm}\subfigure[\normalsize Utility Accuracy]{\includegraphics[width=0.245\textwidth]{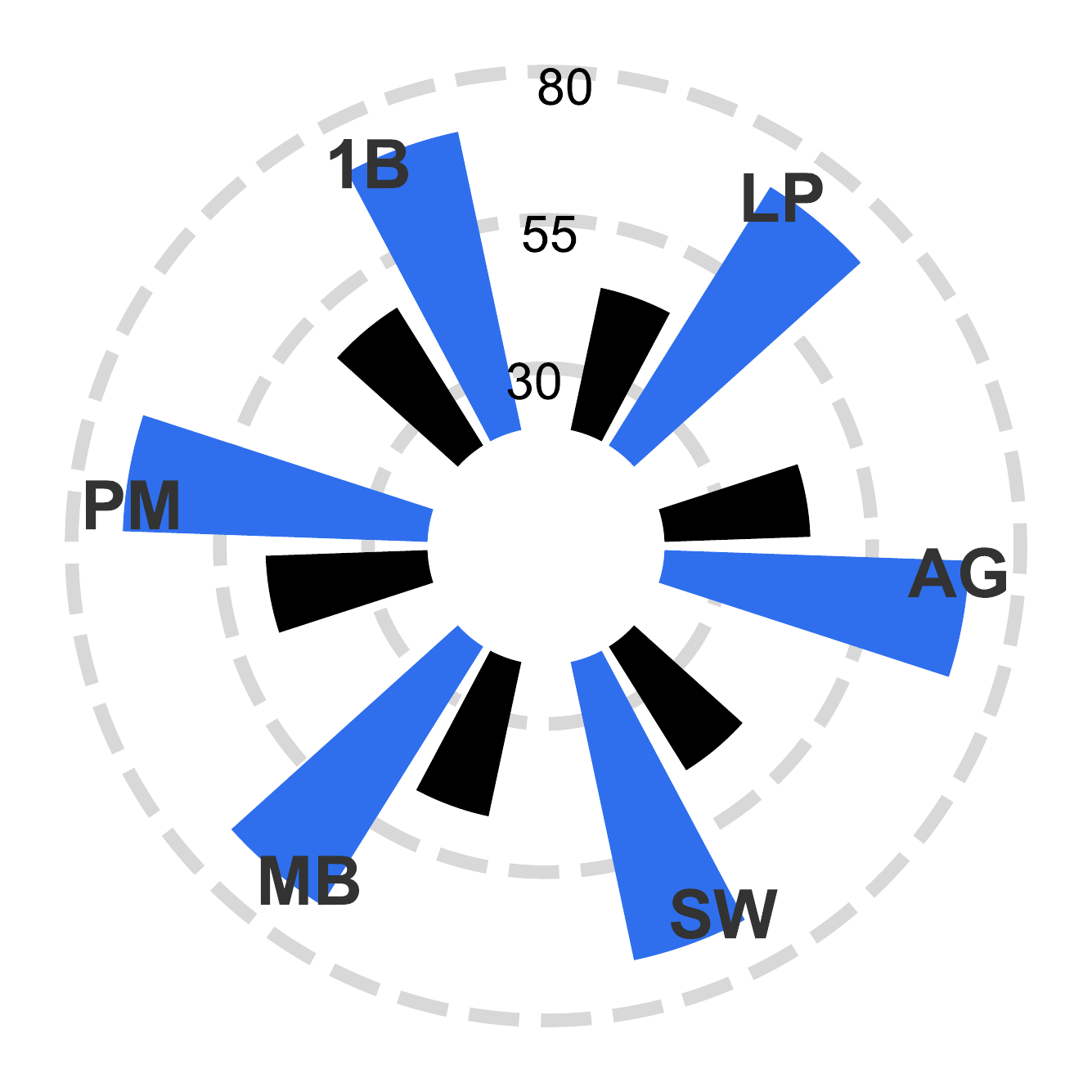}}
     \end{minipage}\\ 
     \vspace{-1em}
     
   \caption{Effectiveness of six local embedding DP mechanisms under a fixed privacy budget $\epsilon_e = 5.0$ on Tape-Arxiv23 with GCN and Linq; \textcolor{blue}{blue} bars denote results before perturbation. (a)-(c) show attack performance for \texttt{MLA}, \texttt{NIA}, and \texttt{MIA}, respectively. (d) reports the corresponding model utility.}

    \label{fig88}
    \vspace{-1em}
\end{figure*}

\vspace{-1em}
\subsection{Vulnerability to Label Inference}\label{sec5.4}
\vspace{-0.5em}
We evaluate vulnerability to label privacy inference using two representative attacks (\texttt{HGA} and \texttt{NIA}), with results shown in Fig.~\ref{fig:label_main} and additional details provided in Appendix. The results reveal two key observations: \ding{172} LLM-enhanced GNNs consistently exhibit stronger label leakage (measured by attack accuracy (\%)) compared to the \textit{Shallow} baseline, suggesting that incorporating semantic knowledge enables label-related signals to be encoded more explicitly in node representations. \ding{173} Performance differences across LLM-enhanced GNN variants remain relatively moderate, indicating that the elevated vulnerability is not driven by any specific model architecture, but rather reflects a common consequence of leveraging semantically enriched features.

\noindent\textbf{In-depth Analysis.} We analyze the distribution of embedding distances between \textit{intra-class} (same-label) and \textit{inter-class} (different-label) node pairs on the Tape-Arxiv23 dataset using GCN model. The results, presented as cumulative distribution functions (CDFs) in Fig.~\ref{fig7}\textcolor{cyan}{(b)}, show that compared to \textit{Shallow}, \textit{Linq} exhibits a more pronounced distributional gap between intra-class and inter-class distances, indicating a clearer class-wise separation in the embedding space. This suggests that LLM-enhanced GNNs learn a more structured, label-aware representation, whose enhanced discriminability, while improving classification utility, simultaneously makes label inference attacks more effective.

\vspace{-1em}
\subsection{Vulnerability to Membership Inference}\label{sec5.5}
\vspace{-0.5em}
We evaluate membership privacy vulnerability using two representative attacks, namely \texttt{MIA} (measured by AUC (\%)) and \texttt{NMA} (measured by attack accuracy (\%)). Results are reported in Fig.~\ref{fig:membership_main}, with additional analyses provided in Appendix. The results show: \ding{172} LLM-enhanced GNNs consistently exhibit higher membership leakage than the \textit{Shallow} baseline across datasets, suggesting that semantic enrichment amplifies the model's tendency to memorize training-specific patterns and leaves stronger membership signals in the learned representations. \ding{173} Among different LLM-enhanced GNN variants, the degree of membership leakage above the \textit{Shallow} baseline varies across methods, reflecting that different feature enhancers encode training membership signals to different extents.

\noindent\textbf{In-depth Analysis.} Membership inference attacks fundamentally exploit the behavioral discrepancy between member and non-member samples. To investigate this, we compare the prediction confidence distributions of member and non-member nodes on Tape-Arxiv23 using GCN, as shown in Fig.~\ref{fig7}\textcolor{cyan}{(c)}. Compared to \textit{Shallow}, \textit{Linq} exhibits a more pronounced distributional gap between member and non-member confidence scores, indicating that LLM-enhanced representations encode stronger training-specific signals. This makes the behavioral discrepancy between member and non-member nodes more distinguishable, thereby amplifying the risk of membership inference attacks.

\vspace{-1em}

\subsection{Effectiveness of Defenses}\label{sec5.6}

\vspace{-0.5em}

In this section, we evaluate representative DP defenses under both input-level perturbation and private GNN training settings. The main-text experiments focus on embedding-, link-, and label-level LDP on Tape-Arxiv23 using GCN with Linq as the feature enhancer. We further evaluate GAP-based private GNN training and cross-backbone defense generalization in Appendix.

\begin{figure*}[t]
    \centering
    \begin{minipage}{\textwidth}
        \centering
        \subfigure[\normalsize Embedding-level DP]{\includegraphics[width=0.32\textwidth]{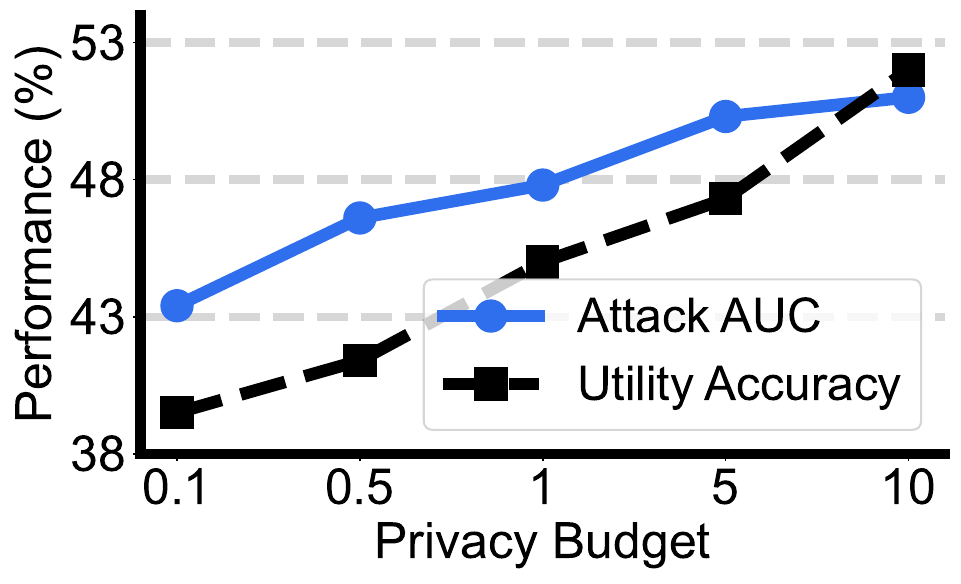}}
        \hspace{0.15cm}\subfigure[\normalsize Link-level DP]{\includegraphics[width=0.32\textwidth]{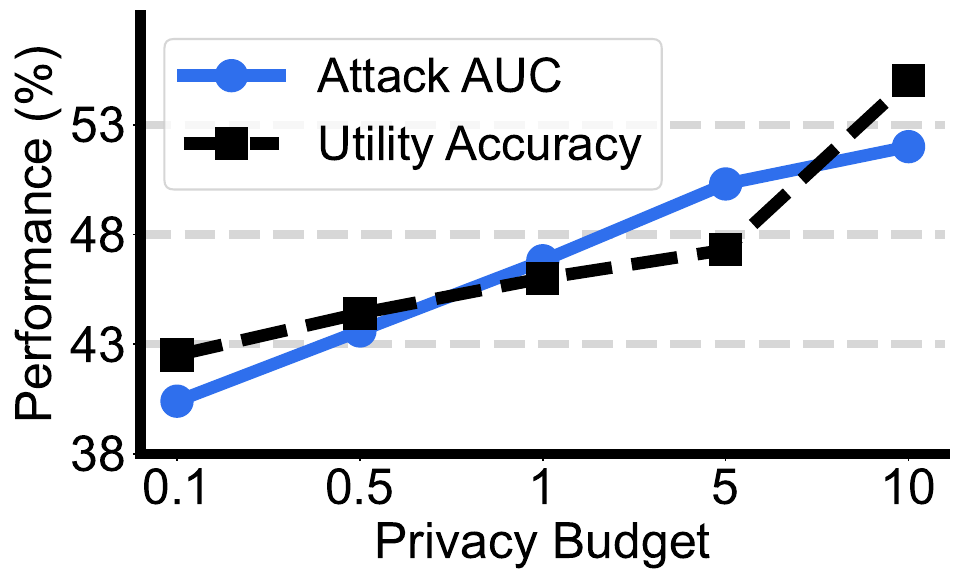}}
        \hspace{0.15cm}\subfigure[\normalsize Label-level DP]{\includegraphics[width=0.32\textwidth]{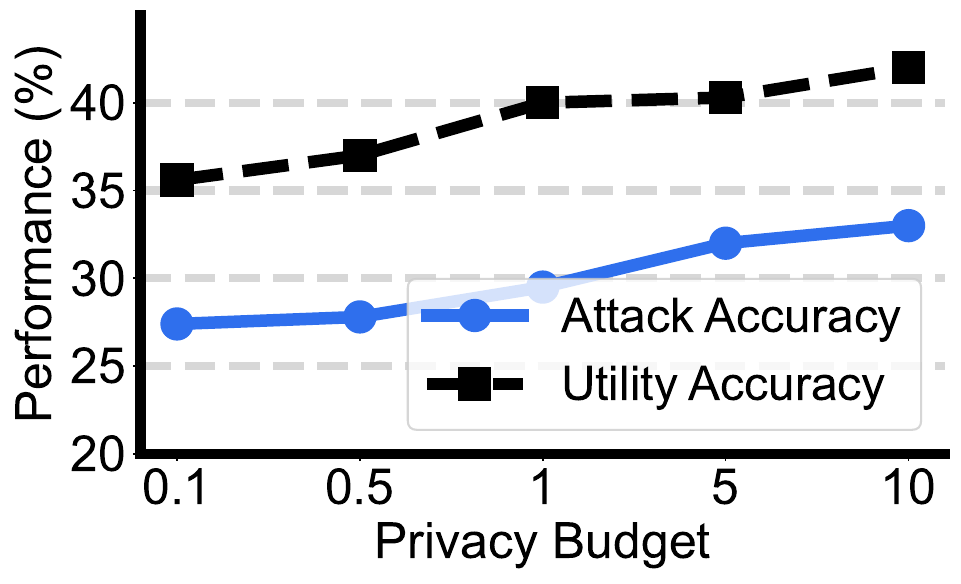}}
     \end{minipage}\\ 
     \vspace{-1em}
  \caption{Attack and utility performance under varying privacy budgets on Tape-Arxiv23 with GCN and Linq. (a) Embedding-level DP ($\epsilon_e$) \textit{vs.}\ \texttt{MLA} (AUC \%) and utility (Accuracy \%). (b) Link-level DP (RR) \textit{vs.}\ \texttt{MLA} (AUC \%) and utility. (c) Label-level DP ($k$-RR) \textit{vs.}\ \texttt{HGA} (Accuracy \%) and utility.}

    \label{fig99}
    \vspace{-1.5em}
\end{figure*}

\noindent\textbf{Embedding DP.} We consider six local DP mechanisms for node embedding perturbation (1B, LP, AG, SW, MB, PM), and evaluate their effectiveness against \texttt{MLA}, \texttt{NIA}, and \texttt{MIA} for link, label, and membership inference, respectively, on Tape-Arxiv23 using GCN with Linq as the feature enhancer. The results are shown in Fig.~\ref{fig88}. Under a moderate privacy budget ($\epsilon_e = 5.0$), Fig.~\ref{fig88}\textcolor{cyan}{(a)-(c)} show that embedding perturbation can mitigate all three types of attacks to varying degrees. However, Fig.~\ref{fig88}\textcolor{cyan}{(d)} reveals a significant drop in model utility, indicating a clear privacy-utility trade-off. We further examine the effect of varying the privacy budget $\epsilon_e \in \{0.1, 0.5, 1.0, 5.0, 10.0\}$ (under the MB) on \texttt{MLA} (Fig.~\ref{fig99}\textcolor{cyan}{(a)}). As $\epsilon_e$ decreases, the performance of the fixed MLA attack and model utility both decline, indicating a privacy-utility trade-off under the evaluated attack protocol.

\noindent\textbf{Link DP.} We further study link-level local DP using the RR. As shown in Fig.~\ref{fig99}\textcolor{cyan}{(b)}, applying RR to the graph structure reduces the effectiveness of \texttt{MLA}. Nevertheless, it simultaneously incurs noticeable utility degradation,  indicating a poor privacy-utility trade-off for structure-level protection.

\noindent\textbf{Label DP.} We examine label-level DP using the $k$-RR mechanism against \texttt{HGA}-based label inference. As shown in Fig.~\ref{fig99}\textcolor{cyan}{(c)}, while label perturbation partially reduces attack performance, it inevitably weakens the learning of true label distributions, leading to degraded classification accuracy. This highlights the inherent tension between protecting label privacy and maintaining model utility.

\vspace{-1em}
\section{Related Work}

\vspace{-0.5em}

In this section, we provide a brief overview of \textit{LLM-enhanced GNNs} and \textit{privacy attacks on GNNs}. A more comprehensive discussion and additional references are provided in Appendix.

\textbf{LLM-Enhanced GNNs.} Recent studies have explored integrating LLMs with GNNs to encode rich textual attributes, giving rise to \textit{LLM-enhanced GNNs}~\citep{DBLP:conf/iclr/HeB0PLH24,chen2024exploring,zhu2024efficient,DBLP:conf/iclr/ChienCHYZMD22,zhang2025toward}, which significantly improve node representations and model utility. Existing methods fall into two categories: \textit{explanation-based methods}, which use LLMs to generate semantic descriptions that are subsequently encoded by a language model~\citep{DBLP:conf/iclr/HeB0PLH24}; and \textit{embedding-based methods}, which directly encode textual attributes via LLMs or fine-tuned language models~\citep{duan2023simteg}. While these approaches have demonstrated significant utility gains, the privacy implications of incorporating semantically enriched representations into graph learning remain largely unexplored, motivating our systematic investigation.

\textbf{Privacy Attacks on GNNs.} Privacy risks in graph learning are commonly studied under three threat scenarios. Membership inference attacks~\citep{olatunji2021membership,guan2025attention,he2021node} determine whether a node was used for training. Link inference attacks~\citep{wu2022linkteller,he2021stealing} recover sensitive graph structure from node representations. Label inference attacks~\citep{meng2023devil} infer private node labels. Prior work mainly studies these threats in conventional GNNs, while their behavior under semantically enriched LLM-based representations remains underexplored.

\vspace{-1em}
\section{Conclusion}
\vspace{-0.5em}
In this work, we systematically audit privacy risks in LLM-enhanced GNNs across link, label, and membership inference. Extensive experiments across ten TAG datasets and diverse model configurations show that LLM-enhanced GNNs generally exhibit greater empirical privacy vulnerability than shallow baselines, accompanied by more distinguishable privacy-sensitive signals. We further show that DP can mitigate these risks, but often at the cost of model utility. Overall, our findings provide practical insights for developing more secure and trustworthy graph learning systems.

\bibliography{iclr2027_conference}
\bibliographystyle{iclr2027_conference}

\end{document}